\documentclass{new_tlp}
\usepackage{mathptmx}
\usepackage[mathcal]{euscript}

\usepackage{amsmath,amssymb}
\usepackage{graphicx}
\graphicspath{ {./} }
\usepackage{caption}
\usepackage{subcaption}
\usepackage{xspace}
\usepackage{url}
\usepackage[hidelinks]{hyperref}

\usepackage[ruled,vlined,noend]{algorithm2e}

\usepackage[normalem]{ulem}
\usepackage{xcolor}

\newcommand\bcmdtab{\noindent\bgroup\tabcolsep=0pt%
  \begin{tabular}{@{}p{10pc}@{}p{20pc}@{}}}
\newcommand\ecmdtab{\end{tabular}\egroup}

\newtheorem{example}{Example}[section]

\newcommand{\Or}{\ensuremath{\ |\ }\xspace}

\newcommand{\naf}{\ensuremath{not\ }\xspace}

\newcommand{\dlv}{\textsc{dlv}\xspace}
\newcommand{\idlv}{\textsc{i-dlv}\xspace}

\newcommand{\dlvdueserver}{\textsc{dlv}{\small{2}}-\textsc{server}\xspace}
\newcommand{\owltodlv}{\textsc{owl}{\small{2}}\textsc{dlv}\xspace}

\newcommand{\hornshiq}{Horn-$\mathcal{SHIQ}$}

\newcommand{\sysname}{{\em DaRLing}\xspace}

\title[DaRLing: A Datalog rewriter for OWL 2 RL ontological reasoning]
{DaRLing: A Datalog rewriter for OWL 2 RL ontological reasoning under SPARQL queries\thanks{This work has been partially supported by MISE under the project ``S2BDW'' (F/050389/01-03/X32) -- Horizon 2020 PON I\&C 2014-2020 and by Regione Calabria under the project ``DLV LargeScale'' (CUP J28C17000220006) -- POR Calabria 2014-2020.}}

  \author[A. Fiorentino, J. Zangari and M. Manna]
  {
    ALESSIO FIORENTINO, 
    JESSICA ZANGARI 
    and MARCO MANNA 
    \\
    Department of Mathematics and Computer Science (DeMaCS), University of Calabria, Rende, Italy
    \\
    \email{{\em lastname}@mat.unical.it} - \url{https://www.mat.unical.it}
  }

\jdate{}
\pubyear{}
\submitted{Jun 2020}
\revised{Jul 2020}
\accepted{Jul 2020}

\begin{document}

\label{firstpage}

\maketitle

\begin{abstract}
The W3C Web Ontology Language (OWL) is a powerful knowledge representation formalism at the basis of many semantic-centric applications. Since its unrestricted usage makes reasoning undecidable already in case of very simple tasks, expressive yet decidable fragments have been identified. Among them, we focus on OWL 2 RL, which offers a rich variety of semantic constructors, apart from supporting all RDFS datatypes. Although popular Web resources - such as DBpedia - fall in OWL 2 RL, only a few systems have been designed and implemented for this fragment. None of them, however, fully satisfy all the following desiderata: $(i)$ being freely available and regularly maintained; $(ii)$ supporting query answering and SPARQL queries; $(iii)$ properly applying the {\em sameAs} property without adopting the unique name assumption; $(iv)$ dealing  with concrete datatypes.
To fill the gap, we present {\em DaRLing},
a freely available Datalog rewriter for OWL 2 RL ontological reasoning under SPARQL queries.
In particular, we describe its architecture, the rewriting strategies it implements, and the result of an experimental evaluation that demonstrates its practical applicability.
This paper is under consideration in Theory and Practice of Logic Programming (TPLP).
\end{abstract}


\section{Introduction}
{\em Ontology-mediated query answering} is an emerging paradigm at the basis of many semantic-centric applications~\cite{DBLP:conf/ijcai/Bienvenu16}. In this setting, a classical data source is reinterpreted via an ontology, which provides a semantic conceptual view of the data. As a direct and positive effect, the knowledge provided by the ontology can be used to improve query answering. Among the formalisms that are capable to express such a conceptual layer, the {\em Web Ontology Language} (OWL) is certainly the most popular one~\cite{owl-web,owl2-overview}. But an ontology-mediated query (OMQ) has also a second component other than the ontology: the actual query that specifies, in a semantic way and via the ontological vocabulary, which part of the data one is interested in. And the most suitable formalism used to specify a query that complements an OWL ontology is definitely the {\em SPARQL Protocol and RDF Query Language} (SPARQL), representing -- as for OWL -- a W3C standard~\cite{harris2013sparql}.

OWL is a very powerful formalism. But its unrestricted usage makes reasoning undecidable already in case of very simple tasks such as fact entailment.
Hence, expressive yet decidable fragments have been identified. Among them, we focus in this paper on the one called OWL 2 RL~\cite{owl2-profiles}. From the knowledge representation point of view, OWL 2 RL enables scalable reasoning without scarifying too much the expressiveness.
Indeed, it supports all RDFS datatypes and provides a rich variety of semantic constructors, such as: {\em inverseOf}, {\em transitiveProperty}, {\em reflexiveProperty}, {\em equivalentClass}, {\em disjointWith}, {\em unionOf},  {\em minCardinality}, {\em allValuesFrom}, {\em someValuesFrom}, and {\em sameAs} -- among others. But the simple fact of allowing {\em someValuesFrom} only in the left-hand-side of an axiom guarantees that conjunctive query answering can be performed in polynomial time in data complexity (when the OMQ is considered fixed) and in nondeterministic polynomial time in the general case (the latter being exactly the same computational complexity of evaluating a single conjunctive query over a relational database).

Although a number of important Web semantic resources -- such as DBpedia%
\footnote{See \url{https://wiki.dbpedia.org/}}
and FOAF%
\footnote{See \url{http://www.foaf-project.org/}} -- trivially fall in OWL 2 RL, only a few systems have been designed and implemented in this setting.
None of them, however, fully satisfy all the following desiderata: $(i)$ being freely available and regularly maintained; $(ii)$ supporting ontology-mediated query answering; $(iii)$ properly applying the {\em sameAs} property without adopting the unique name assumption; $(iv)$ dealing  with concrete {\em datatypes}.

\begin{table}
	\caption{Main tools for ontology-mediated query answering over OWL 2 RL knowledge bases.}
	\label{tab:systems_features}
	\begin{minipage}{\textwidth}
		\centering
		\begin{tabular}{cccccc}
			\hline\hline
			\multicolumn{1}{c}{{Tool}} &
			\multicolumn{1}{c}{{License}} & \multicolumn{1}{c}{{~Latest release~}} & \multicolumn{1}{c}{{~Query language~}} & \multicolumn{1}{c}{{ ~owl:sameAs~}} &
			\multicolumn{1}{c}{{~Datatypes~}} \\
			\hline
			    \multicolumn{1}{c}{Clipper} &
			    \multicolumn{1}{c}{Free} &
			    \multicolumn{1}{c}{Dec 2015} &
			    \multicolumn{1}{c}{SPARQL-BGP} &
			    \multicolumn{1}{c}{under UNA} &
			    \multicolumn{1}{c}{No}
			    \\
			    \multicolumn{1}{c}{\sysname} &
			    \multicolumn{1}{c}{Free} &
			    \multicolumn{1}{c}{Jul 2020} &
			    \multicolumn{1}{c}{SPARQL-BGP} &
			    \multicolumn{1}{c}{Yes} &
			    \multicolumn{1}{c}{Yes}
			    \\
			    \multicolumn{1}{c}{DReW} &
			    \multicolumn{1}{c}{Free} &
			    \multicolumn{1}{c}{Mar 2013} &
			    \multicolumn{1}{c}{SPARQL-BGP} &
			    \multicolumn{1}{c}{No} &
			    \multicolumn{1}{c}{No}
			    \\
			    \multicolumn{1}{c}{Orel} &
			    \multicolumn{1}{c}{Free} &
			    \multicolumn{1}{c}{Feb 2010} &
			    \multicolumn{1}{c}{ground queries} &
			    \multicolumn{1}{c}{No} &
			    \multicolumn{1}{c}{No}
			    \\
			    \multicolumn{1}{c}{\owltodlv} &
			    \multicolumn{1}{c}{Commercial} &
			    \multicolumn{1}{c}{Jun 2019} &
			    \multicolumn{1}{c}{SPARQL-BGP} &
			    \multicolumn{1}{c}{under UNA} &
			    \multicolumn{1}{c}{Yes}
			    \\
			    \multicolumn{1}{c}{~OwlOntDB~} &
			    \multicolumn{1}{c}{-} &
			    \multicolumn{1}{c}{-} &
			    \multicolumn{1}{c}{SPARQL-DL$_E$} &
			    \multicolumn{1}{c}{under UNA} &
			    \multicolumn{1}{c}{No}
			    \\
			    \multicolumn{1}{c}{RDFox} &
			    \multicolumn{1}{c}{~~Commercial~~} &
			    \multicolumn{1}{c}{Jun 2020} &
			    \multicolumn{1}{c}{SPARQL 1.1} &
			    \multicolumn{1}{c}{Yes} &
			    \multicolumn{1}{c}{Yes} \\
			\hline\hline
		\end{tabular}
		\end{minipage}
	\end{table}

To fill this gap, we present {\em DaRLing}\footnote{See \url{https://demacs-unical.github.io/DaRLing/}},
a freely available Datalog rewriter for OWL 2 RL ontological reasoning under SPARQL queries.
Table~\ref{tab:systems_features} reports the main tools  supporting or implementing natively ontology-mediated query answering over knowledge bases that fall in the RL profile of OWL 2, or beyond.
Concerning the query language, apart from Orel, all the tools support SPARQL patterns:
SPARQL 1.1,
SPARQL-BGP~\cite{harris2013sparql}, and
SPARQL-DL$_E$~\cite{DBLP:conf/owled/SirinP07}.
Finally, the row of OwlOntDB contains some missing value because the system  is currently not available.
Hence, none of the existing systems fully meet conditions $(i)$-$(iv)$ above.
(A deeper comparison and discussion is reported in Section~\ref{sec:related}.)

\sysname can take in input an RDF dataset (ABox) $\mathcal{A}$, an OWL 2 RL ontology (TBox) $\mathcal{T}$ and a SPARQL query $q({\bf x})$, and constructs an equivalent program $P$ with an output predicate $\mathit{ans}$ of arity $|{\bf x}|$. Formally, for each $|{\bf x}|$-tuple of domain constants, $\mathcal{A} \cup \mathcal{T} \models q({\bf t})$ if, and only if, the atom $\mathit{ans}({\bf t})$ can be derived via $P$, where $P$ is a Datalog program using inequality and stratified negation.
The system builds on top of the OWL API. It supports different input formats and knowledge bases organized in multiple files. Moreover, it can produce a suitable rewriting also if some inputs are missing. For example, in case the ABox is missing, then the generated program is simply equivalent to the pair TBox plus query.

In what follows, after some background material (Section~\ref{sec:Background}),
we describe {\em DaRLing}'s architecture (Section~\ref{sec:system}) and the rewriting strategies it implements (Sections~\ref{sec:translation} and~\ref{sec:sameAs}). We discuss related work (Section~\ref{sec:related}) and eventually we report the result of an experimental evaluation (Section~\ref{sec:eval}) before drawing some conclusions (Section~\ref{sec:conclusion}).

\section{Background}\label{sec:Background}
We assume that the reader is familiar with the basic notions about Datalog~\cite{DBLP:journals/tkde/CeriGT89} and Description Logics~\cite{baader2008description}.
In this section we provide the syntactic notation used in the remainder of the paper.

\subsection{Datalog}
\label{subsubsec:datalog}
A {\em term} is either a constant or a variable. If $t_1,\dots, t_k$ are terms and $p$ is a \emph{predicate symbol} of arity $k$, then $p(t_1,\dots, t_k)$ is a {\em predicate atom}. A {\em built-in atom} has form $t \prec u$, where $t$ and $u$ are terms, and \mbox{$\prec\ \in \{``<"\,``\leq",``>",``\geq",``=",``\neq"\}$.}
A {\em classical literal} $\,l$ is of the form $a$ or $\naf a$, where $a$ is a predicate atom; in the former case $l$ is {\em positive}, otherwise {\em negative}. If $l$ is a classical literal or a built-in atom, then $l$ is called {\em naf-literal}.
A {\em rule $r$} is of the form
$$ h\ \gets\ b_1\ \land\ \dots\ \land\ b_n\, .$$
where $n\geq 0$; $h$ (the {\em head} of $r$) is a predicate atom, and $b_1, \dots, b_{n}$ (the {\em body literals} of $r$) are naf-literals.
A rule $r$ is {\em safe} if each variable has an occurrence in at least a positive classical body literal of $r$.
A {\em program} is a finite set $P$ of safe rules stratified with respect to negation.
A program, rule, or literal is {\em ground} if it contains no variables. A ground rule with an empty body is a {\em fact}.
%

\subsection{The OWL 2 RL profile}
\label{sec:descriptionLogics}

Description logics (DLs) are a family of formal knowledge representation languages used to describe and reason about the ``concepts'' of an application domain.
Among the most relevant applications of description logics there is the OWL Web Ontology Language\footnote{See \url{http://www.w3.org/TR/owl2-overview/}}, a knowledge representation language standardised by the World Wide Web Consortium (W3C) and designed to facilitate the development of Semantic Web applications.
The current version of the OWL specification\footnote{See \url{https://www.w3.org/TR/owl2-syntax/}} is OWL 2~\cite{DBLP:journals/ws/GrauHMPPS08}, developed by the W3C OWL Working Group.
The syntactic elements of OWL 2 are almost analogous to those of a DL, with
the main difference that {\em concepts} and {\em roles} are called {\em classes} and {\em properties} respectively.
OWL 2 Profiles (i.e., OWL 2 EL, OWL 2 QL, and OWL 2 RL)\footnote{See \url{http://www.w3.org/TR/owl2-profiles/}} are syntactic restrictions of OWL 2 that offer significant benefits from a computational point of view at the expense of the expressive power.
In the following we provide the notation related to the DL underlying the OWL 2 RL profile.

Let $N_C$ ({\em atomic concepts}), $N_R$ ({\em role names}), and $N_I$ ({\em individuals}) be mutually disjoint discrete sets.
A \emph{role} is either $r\in N_R$ or an inverse role $r^-$ with $r\in N_R$. We denote by $R^-$ the \emph{inverse} of a role $R$ defined by $R^-:=r^-$ when $R=r$ and $R^-:=r$ when $R=r^-$. The set of \emph{concepts} is the smallest set such that: $(i)$ $\top$, $\bot$, and every atomic concept $A\in N_C$ is a concept; $(ii)$ if $C$ and $D$ are concepts and $R$ is a role, then $C\sqcap D$, $C\sqcup D$, $\neg C$, $\forall R.C$, $\exists R.C$, $\geq nR.C$ and $\leq nR.C$, for $n\geq 1$, are concepts.

{\em Subconcept} and {\em Superconcept Expressions} are recursively defined as follows: $(i)$ $\top$ and every atomic concept $A\in N_C$ is both a Subconcept and a Superconcept Expression; ($ii$) if $C$ and $D$ are Subconcept Expressions and $R$ is a role, then $C\sqcap D$, $C\sqcup D$, $\exists R.C$ and $\geq 1R.C$ are Subconcept Expressions; ($iii$) if $C$ and $D$ are Superconcept Expressions and $R$ is a role, then $C\sqcap D$, $\neg C$, $\forall R.C$, and $\leq 1R.C$ are Superconcept Expressions.

A {\em knowledge base} (KB) is any pair $\cal K=(A,T)$ where:
\begin{itemize}
    \item
    [$(i)$] $\cal A$, the ABox ({\em assertional box}), is a finite set
of {\em assertions} of the form $A(a)$ or $R(a, b)$, with $a, b \in N_I$, $A \in N_C$, and $R \in N_R$;
    \item
    [$(ii)$] $\cal T$ , the TBox ({\em terminological box}), is a finite set of: $(a)$ {\em concept inclusions} (CIs) of the form $C\sqsubseteq D$, where $C$ and $D$ are concepts; $(b)$ {\em role inclusions} (RIs) of the form $R\sqsubseteq S$, where $R$ and $S$ are roles; and $(c)$ {\em transitivity axioms} of the form trans$(R)$, where $R$ is a role.
\end{itemize}
We will generally refer to the elements of an ABox or a TBox by calling them {\em axioms}.
We say that a TBox belongs to the OWL 2 RL profile if for each CI $C\sqsubseteq D$ it turns out that $C$ is a Subconcept Expression and $D$ is a Superconcept Expression; a KB belongs to the OWL 2 RL profile if its TBox belongs to it.

\begin{example}
\label{example:non-RL-TBox}
	The TBox
	\begin{align*}
		& \mathtt{linkedViaTrain} \sqsubseteq \mathtt{linked} && \text{trans{\tt (linkedViaTrain)}}\\
		&\mathtt{CommutingArea} \sqsubseteq \exists \mathtt{linked.Capital} &&\exists \mathtt{linked.Capital} \sqsubseteq \mathtt{DesirableArea}  \\
		&\mathtt{Capital} \sqsubseteq \mathtt{DesirableArea}
	\end{align*}
	does not fall in the OWL 2 RL profile.
	In fact, in $\mathtt{CommutingArea}  \sqsubseteq  \exists \mathtt{linked.Capital}$, the concept $\exists{\tt linked.Capital}$ is not a Superconcept Expression. \hfill$\square$
\end{example}

{\em Positive} and {\em negative occurrences} of a concept $C$ in concepts are defined as follows:
\begin{itemize}
    \item
    $C$ occurs positively in itself;
    \item
    $C$ occurs positively (resp., negatively) in $\neg C_-$ or $\leq nS.C_-$ if $C$ occurs negatively (resp., positively) in $C_-\,$; and
    \item
    $C$ occurs positively (resp., negatively) in $C_+ \sqcap D_+$, $C_+ \sqcup D_+$, $\exists R.C_+$, $\forall R.C_+$ or $\geq nS.C_+$ if $C$ occurs positively (resp., negatively) in $C_+$ or in $D_+\,$.
\end{itemize}
We say that a concept $C$ occurs positively (resp., negatively) in an axiom $C_-  \sqsubseteq C_+$ if $C$ occurs positively (resp., negatively) in $C_+$, or negatively (resp., positively) in $C_-$;
$C$ occurs positively (resp., negatively) in $\cal T$ if $C$ occurs positively (resp., negatively) in some axiom of $\cal T$.

\section{The \sysname\ rewriter: system overview}
\label{sec:system}
\begin{figure}[h]
    \centering
    \includegraphics[width=\textwidth]{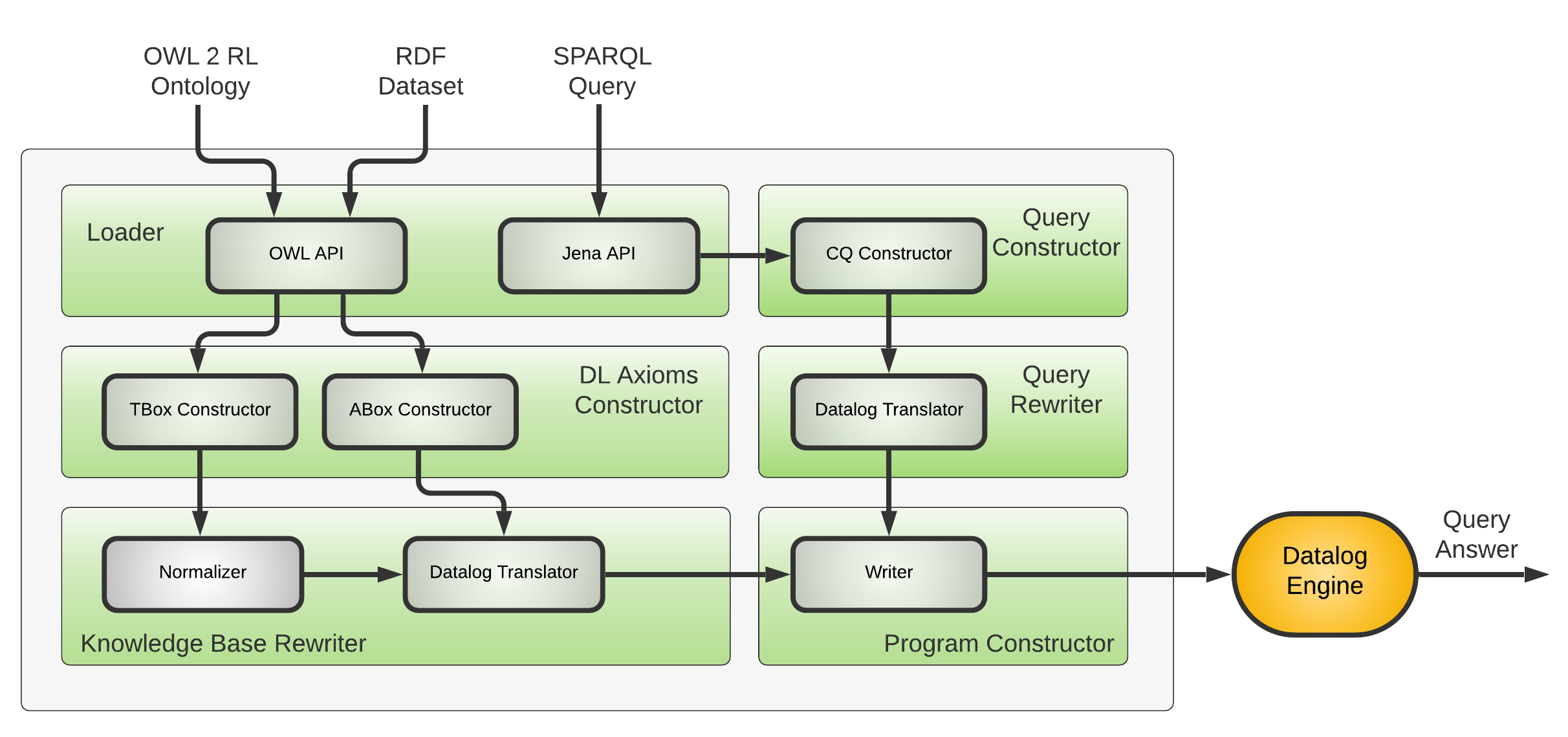}
    \caption{\sysname\ Architecture}
    \label{fig:architecture}
\end{figure}

\sysname\ is an open-source Datalog rewriter for OWL 2 RL ontologies available on the online webpage \url{https://demacs-unical.github.io/DaRLing/}.
The rewriter implements the translation techniques described in Section~\ref{sec:translation} and supports the {\tt owl:sameAs} management method described in Section~\ref{sec:sameAs}.
In addition to the Datalog rewriting of OWL 2 RL ontologies, \sysname\ also supports the Datalog translation of datasets in RDF/XML or Turtle format and SPARQL queries containing only basic graph patterns.
The architecture of \sysname\ is synthesized in Figure~\ref{fig:architecture}.

\sysname\ uses the OWL API~\cite{DBLP:conf/semweb/HorridgeB08} to load the RL fragment of OWL 2 ontologies and datasets into internal data structures representing DL TBoxes and ABoxes, respectively.
The system supports the loading of the datatypes {\em xsd:string} and {\em xsd:integer}, provided by the OWL 2 datatype map (a list of the datatypes that can be used in OWL 2 ontologies) for the representation of strings and integers.
A rewrite module is implemented for the translation of a TBox and/or an ABox into a Datalog program. More in detail, a part of the axioms (those ``directly rewritable'') is passed to the {\em Datalog Translator}, whereas the remaining part is first subjected to the normalization procedure described in Section~\ref{sec:translation}.

\sysname\ also allows for Datalog translation of SPARQL queries containing {\em basic graph patterns} (BGPs) (i.e., sets of triple patterns forming a graph).
SPARQL queries are parsed using Jena~\footnote{See \url{https://jena.apache.org/}} -- a Java API which can be used to create and manipulate RDF graphs -- then they are translated into conjunctive Datalog queries.

The system supports different input formats and knowledge bases organized in multiple files.
More in detail:
$(i)$ the ontology (resp., the dataset) can be contained in a single file or in a folder containing multiple files with one of the extensions {\em .owl} or {\em .rdfs} (resp., {\em .owl}, {\em .rdf} or {\em .ttl});
$(ii)$ one or more queries have to be contained in a file with the {\em .SPARQL} extension.
Moreover, \sysname\ can produce a suitable rewriting also if some inputs are missing: for each file (or folder) received as input, a single {\em .asp} file containing the respective Datalog translation is returned as output. For example, in case the ABox is missing, then the generated program is simply equivalent to the pair TBox plus query.

By default, \sysname\ rewrites under the {\em Unique Name Assumption} (UNA).
However, it is possible to explicitly choose to enable rewriting with the {\tt owl:sameAs} management mode described in Section~\ref{sec:sameAs} if at least one between ontology and query is given as input.
The semantics of {\tt owl:sameAs}, indeed, presuppose the enabling of matches between syntactically different but equivalent individuals. Therefore, its management is strictly linked to its propagation among the join variables of the Datalog rewriting of the ontology or the query (cfr. Section~\ref{sec:sameAs}).
%

\section{From OWL 2 RL to Datalog}
\label{sec:translation}

In this section we describe how the TBox underlying an OWL 2 RL ontology is rewritten into Datalog.
Since translating RIs and transitive axioms is almost trivial, we will focus only on the CIs.
In particular, the rewriting process takes place as follows: $(i)$ a class of axioms for which we provide a direct translation is identified; $(ii)$ the remaining part of the axioms is subjected to a normalization procedure before being translated.
%

\subsection{A direct translation for a class of CIs}\label{subsec:algorithm}

For the first step we need to introduce the concepts of the ${\cal ELI}$ description logic.
{\em ${\cal ELI}$-concepts} are inductively defined as follows: $(a)$ $\top$ and each $A \in N_C$ is an ${\cal ELI}$-concept, and $(b)$ if $C$, $D$ are ${\cal ELI}$-concepts and $R$ is a role, then $C \sqcap D$, $\exists R.C$ and $\geq 1 R.C$ are ${\cal ELI}$-concepts.
We show how to get the equivalent Datalog rule for an axiom of the form \ $\sqcup C_i \sqsubseteq \sqcap A_i \,$, where $\sqcup C_i$ is a disjunction of ${\cal ELI}$-concepts and $\sqcap A_i$ is a conjunction of atomic concepts.
Since $\sqcup C_i \sqsubseteq \sqcap A_i $ is equivalent to the axioms $C_j \sqsubseteq A_k \,$ for each $C_j$ and $A_k$, it is sufficient to show how the translation works on CIs of the form $C \sqsubseteq A$, where $C$ is an ${\cal ELI}$-concept and $A$ is atomic.
Algorithm~\ref{algorithm:ELIconcepts} shows a recursive algorithm for generating literals starting from an ${\cal ELI}$-concept $C$. Intuitively, for each conjunction $\sqcap C_i$ (possibly consisting of a single clause), we have a variable common to each clause $C_i$. For every clause $C_i$ of the form $\exists R.D$ or $\geq 1R.D$ we introduce a fresh variable obtained by adding $i$ to the subscript of the variable shared by the conjunction to which $C_i$ belongs.

\begin{algorithm}[]
\SetAlgoLined
\SetKwInOut{Input}{Input}
\Input{ {$\cal ELI$}-concept $C$, String Var, int clause, Set$<$Literal$>$ bodyLiterals}
\KwResult{Addition of Datalog literals to bodyLiterals}

  \uIf{$C$ is atomic}{
   bodyLiterals.add($C$(Var))\;
   }
  \uElseIf{$C$ has form $\sqcap C_i$}{
    \ForEach{i}{
        translate${\cal ELI}$($C_i$, Var, $i$, bodyLiterals)\;
    }
  }
  \uElseIf{$C$ has form $\exists R.D$ \Or $\geq 1R.D$}{
   String newVar = Var + `\_' + clause \;
   \uIf{$R$ is an inverse role}{
       bodyLiterals.add($R^-$(newVar,Var)) \;
   }
   \uElse {
        bodyLiterals.add($R$(Var,newVar))
   }
   translate${\cal ELI}$($D$, newVar, $1$, bodyLiterals)\;
  }

 \caption{translate${\cal ELI}$($C$, Var, clause, bodyLiterals)}
 \label{algorithm:ELIconcepts}
\end{algorithm}
The translation of an axiom $C \sqsubseteq A \,$ (as above) is a rule whose head is the literal $A(X)$ and whose body literals are obtained by invoking Algorithm~\ref{algorithm:ELIconcepts} on $C$ with $\text{Var} = X$ and $\text{clause} = 1$.

\begin{example}
The concept inclusion
$\exists r.(\exists s.(C\sqcap D))\ \sqcap\ \geq 1 t.(E\sqcap \exists u^-.F) \sqsubseteq A$
translates directly to the following Datalog rule:
\begin{align*}
    A(X) \gets r(X,X_1) \land s(X_1,X_{1,1}) \land C(X_{1,1}) \land D(X_{1,1}) &\ \land \\ t(X,X_2) \land E(X_2) \land u(X_{2,1},X_2) \land F(X_{2,1}).
\end{align*}
In more detail, the variable shared by clauses $r.(\exists s.(C\sqcap D))$ and $\geq 1 t.(E\sqcap \exists u^-.F)$ is $X$. The recursive call on these two clauses generates the body literals $\{ r(X,X_1)$, $s(X_1,X_{1,1})$, $C(X_{1,1})$, $D(X_{1,1}) \}$ and $\{ t(X,X_2) $, $E(X_2)$, $u(X_{2,1},X_2)$, $F(X_{2,1}) \}$ respectively.\hfill$\square$
\end{example}

\subsection{Enhancing the normalization procedure}

For the second phase of the rewriting process, we transform the remaining axioms of the TBox (those not directly translatable) into a ``normalized'' form, from which the Datalog translation is immediate.
We say that a TBox ${\cal T}$ is in {\em normalized form} if each concept inclusion axiom in ${\cal T}$ has form\ $\sqcap A_i  \sqsubseteq C$, where $\sqcap A_i$ is a conjunction of atomic concepts and $C$ is a concept of the form $\bot$, $A$, $\forall R.A$ or $\leq 1 R.A$, with $A$ atomic.
Table~\ref{table:translation_of_normalized_axiom} shows how the Datalog translation of a TBox in normalized form takes place; in particular, $A,A_1,\dots,A_n$ are atomic concepts and $R(X,Y)=r(Y,X)$ if $R=r^-$.
\begin{table}[b]
\caption{Translation of Concept Inclusions in normalized form.}
\label{table:translation_of_normalized_axiom}
\begin{minipage}{\textwidth}
    \centering
\begin{tabular}{cc}
    \hline\hline
	\multicolumn{1}{l}{{Concept Inclusion (CI)}}  & \multicolumn{1}{c}{{Equivalent Datalog Rule}}  \\
	\hline
	\multicolumn{1}{l}{$A_1 \sqcap\dots\sqcap A_n  \sqsubseteq \bot$} &  \multicolumn{1}{c}{$\bot\ \gets\ A_1(X)\ \land\ \dots\ \land\ A_n(X).$} \\
	\multicolumn{1}{l}{$A_1 \sqcap\dots\sqcap A_n  \sqsubseteq A$} & \multicolumn{1}{c}{$A(X)\ \gets\ A_1(X)\ \land\ \dots\ \land\ A_n(X).$} \\	
	\multicolumn{1}{l}{$A_1 \sqcap\dots\sqcap A_n  \sqsubseteq \forall R.A$} & \multicolumn{1}{c}{$A(Y)\ \gets\ R(X,Y)\ \land\ A_1(X)\ \land\ \dots\ \land\ A_n(X).$} \\
	\multicolumn{1}{l}{$A_1 \sqcap\dots\sqcap A_n  \sqsubseteq\; \leq 1 R.A$~~~} & \multicolumn{1}{c}{{\it sameAs}$(Y_1,Y_2)\ \gets\ A_1(X)\ \land\ \dots\ \land\ A_n(X)\ \land\ $}\\
	\multicolumn{1}{l}{} & \multicolumn{1}{c}{~~~$R(X,Y_1)\ \land\ R(X,Y_2)\ \land\ A(Y_1)\ \land\ A(Y_2)\ \land\ Y_1\neq Y_2.$~~~}\\
    \hline\hline
\end{tabular}
\vspace{-1\baselineskip}
\end{minipage}
\end{table}
We bring our axioms to a normalized form readapting a normalization procedure described by Kazakov~\cite{DBLP:conf/ijcai/Kazakov09}, which is applicable to any \hornshiq\ TBox and preserves the logical consequences of the ontological axioms.
If on the one hand normalization allows us to easily translate a given ontology into Datalog, on the other, it could significantly increase the number of axioms.
In what follows we describe, with the help of some examples, a version of the normalization procedure ad-hoc for OWL 2 RL.
We also show how we enhance that procedure in order to avoid, where possible, an unnecessary growth of the number of axioms.

Normalization aims at reducing the complex structure of axioms by introducing fresh concept names for substructures and substituting them.
Intuitively, the transformation works as follows: let $C$ be a complex concept containing $D$ as a sub-expression; then, a fresh concept name $A_D$ is introduced and constrained to extensionally coincide with $D$. This enables us to exchange all occurrences of $D$ in $C$ \mbox{by $A_D$}.

Formally, given a OWL 2 RL TBox $\cal T$,  for every (sub-)concept $C$ in $\cal T$ we introduce a fresh atomic concept $A_C$ and define a function $st(C)$ by:
\begin{align*}
	&st(A)=A \text{ ($A$ atomic)}; &&st(\bot)=\bot; &&st(\top)=\top;\\
	&st(\neg C)=\neg A_C;&&st(C\sqcap D)=A_C\sqcap A_D; &&st(C\sqcup D)=A_C\sqcup A_D;\\
	&st(\forall R.C)=\forall R.A_C; &&st(\exists R.C)=\exists R.A_C;&&st(\geq nR.C)=\,\geq nR.A_C; \\
	&st(\leq nR.C)=\,\leq nR.A_C\;.
\end{align*}
The result of applying \emph{structural transformation} to $\cal T$ is an ontology $\cal T'$ that contains all role inclusions and transitivity axioms in $\cal T$ in addition to the following axioms:
\begin{itemize}
	\item $A_C\sqsubseteq st(C)$ for every $C$ occurring positively in $\cal T$
	\item $st(C)\sqsubseteq A_C$ for every $C$ occurring negatively in $\cal T$	
	\item $A_C\sqsubseteq A_D$ for every concept inclusion $C\sqsubseteq D\in \cal T$
\end{itemize}
\begin{example}
\label{example:structural_transformation}
	The axiom $\exists r.(B \sqcap C) \sqsubseteq \forall s^-.D$ will be
	transformed into:
	\begin{align*}
	& (R.1)\hspace{0.25em} A_{\forall s^-.D} \sqsubseteq \forall s^-.A_D
	 && (R.2)\hspace{0.25em} A_D \sqsubseteq D
	 && (R.3)\hspace{0.25em} \exists r.(A_{B \sqcap C}) \sqsubseteq A_{\exists r.(B \sqcap C)} \\
	& (R.4)\hspace{0.25em} A_B \sqcap A_C \sqsubseteq A_{B \sqcap C}
	 && (R.5)\hspace{0.25em} B \sqsubseteq A_B
	 && (R.6)\hspace{0.25em} C \sqsubseteq A_C \\
	& (R.7)\hspace{0.25em} A_{\exists r.(B \sqcap C)} \sqsubseteq A_{\forall s^-.D} && &&
	\end{align*}
	where $(R.1)$-$(R.2)$ derive from the positive occurrences of the concepts $\forall s^-.D$ and $D$, whereas $(R.3)$-$(R.6)$ derive from the negative occurrences of $\exists r.(B \sqcap C)$, $B \sqcap C$, $B$ and $C$, respectively.\hfill$\square$
\end{example}
By applying structural transformation to $\cal T$, we obtain a TBox $\cal T'$ containing only concept inclusions of the form $A_1\sqsubseteq A_2$, $A\sqsubseteq st(C_+)$ and $st(C_-)\sqsubseteq A$, where $C_+$ occurs positively and $C_-$ occurs negatively in $\cal T$. Since $\cal T$ belongs to OWL 2 RL,
$C_+$ can only be of the form $\top$, $A$, $\neg C$, $C\sqcap D$,  $\forall R.C$ or $\leq 1R.C$, whereas
$C_-$ can only be of the form $\top$, $A$, $C\sqcap D$, $C\sqcup D$, $\exists R.C$ or $\geq 1R.C$.
Therefore, axioms in $\cal T'$ which do not appear in normalized form are transformed as follows:
\begin{align*}
	& A\sqsubseteq st(\neg C) = \neg A_C &&\hspace{-2cm}\Longrightarrow A\sqcap A_C\sqsubseteq \bot;\\
	& A\sqsubseteq st(C\sqcap D) = A_C\sqcap A_D &&\hspace{-2cm}\Longrightarrow A\sqsubseteq A_C,\; A\sqsubseteq A_D; \\
	& A_C\sqcup A_D = st(C\sqcup D)\sqsubseteq A &&\hspace{-2cm}\Longrightarrow A_C\sqsubseteq A,\; A_D\sqsubseteq A;\\
	& \exists R.A_C = st(\exists R.C)\sqsubseteq A &&\hspace{-2cm}\Longrightarrow A_C\sqsubseteq \forall R^-.A;\\
	& \geq 1R.A_C = st(\geq 1R.C)\sqsubseteq A &&\hspace{-2cm}\Longrightarrow A_C\sqsubseteq \forall R^-.A.
\end{align*}
\begin{example}
\label{example:apply_normalization}
	The axiom $(R.3)$ of the Example~\ref{example:structural_transformation} is not in normalized form and will be replaced with $A_{B \sqcap C} \sqsubseteq \forall r^{-}.A_{\exists r.(B \sqcap C)}$.\hfill$\square$
\end{example}
The rewriting process takes a huge advantage from the fact that many axioms -- those described in Section~\ref{subsec:algorithm} -- are not subject to the normalization procedure.
We further enhance that procedure by providing that: $(i)$ no fresh concept is introduced for $\top$, $\bot$ and all the atomic concepts in the TBox; $(ii)$ axioms already in normalized form are not subjected to the normalization process.
The following example shows how this significantly reduces the number of rewritten axioms (and consequently the number of Datalog rules that derive from them).
\begin{example}
With the concept inclusion of Example~\ref{example:structural_transformation}, through the application of the structural transformation we produce in addition to $(R.3)$ and $(R.7)$ only the two axioms:
\begin{align*}
	& (R'.1)\hspace{0.25em} A_{\forall s^-.D} \sqsubseteq \forall s^-.D
	\hspace{1.5em} & (R'.2)\hspace{0.25em} B \sqcap C \sqsubseteq A_{B \sqcap C}\,.
\end{align*}
Then $(R.3)$ is replaced as in Example~\ref{example:apply_normalization}.
Eventually, a CI already in normalized form like $A \sqcap B \sqsubseteq \forall r.C$, for which the standard normalization would generate $6$ further axioms, is not subjected to the normalization procedure.\hfill$\square$
\end{example}

\section{Handling {\tt owl:sameAs} via Datalog}
\label{sec:sameAs}

The {\tt sameAs} role derives from the {\tt owl:sameAs} property which is used by many OWL 2 ontologies to declare equalities between resources.
The assertion {\tt sameAs}$(a,b)$ states that the individuals $a$ and $b$ are synonyms, i.e., $a$ can be replaced with $b$ without affecting the meaning of the ontology.
As well as logic programming approaches, Datalog works under the {\em Unique Name Assumption} (UNA), i.e., presumes that different names represent different objects of the world.
With the following example we highlight the need of handling the {\tt owl:sameAs} in order to enable the match of equivalent constants for each join between variables in the body of a rule.
\begin{example}
\label{example:griffin}
Let us consider an ontology featuring the rule
\begin{equation*}
\label{rule:griffin}
    \text{DogOwner}(X) \gets \text{hasPet}(X,Y) \ \land\ \text{Dog}(Y).
\end{equation*}
together with the following set of facts:
\begin{equation*}
    \text{hasPet}(\text{``Peter''},\text{``Brian''}).\hspace{1.0em} \text{Dog}(\text{``BrianGriffin''}).\hspace{1.0em} \text{sameAs}(\text{``Brian''},\text{``BrianGriffin''}).
\end{equation*}
Note how, despite that \text{sameAs}(``Brian'',``BrianGriffin'') has the purpose of making the constants ``Brian'' and ``BrianGriffin'' interchangeable, the fact \text{DogOwner}(``Peter'') is not derived as it should.\hfill$\square$
\end{example}

As mentioned in Section~\ref{sec:system}, in order to allow the rewriting with the non-UNA, at least one between ontology and query must be given as input to the system.
Below we provide a way to handle reasoning on a Datalog program deriving from a Web ontology that contains the {\tt owl:sameAs} property.
The idea is to simulate the reflexivity, the symmetry and the transitivity of the {\tt owl:sameAs} through a fresh binary predicate which connects all the elements of a {\tt owl:sameAs}-clique to a representative (the lexicographic minimum) of that clique.
To this end, given $N\geq 0$, we add the following block of rules to our Datalog program:
\begin{align}
\label{sameAs:complem_rule1}
& \textit{sameAs}(X,Y) \gets \textit{sameAs}(Y,X).\\
\label{sameAs:complem_ruleh1}\tag{$2.1$}
& \textit{noStart}(X_1)\ \gets\ \textit{sameAs}(X_0,X_1)\ \land\ X_0<X_1.\\
\label{sameAs:complem_ruleh2}\tag{$2.2$}
& \textit{noStart}(X_2)\ \gets\  \textit{sameAs}(X_0,X_1)\ \land\  \textit{sameAs}(X_1,X_2) \land\ X_0 < X_2.\\
\tag*{\vdots}
& \vdots\\
\label{sameAs:complem_rulehn}\tag{$2.N$}
& \textit{noStart}(X_N)\ \gets\ \textit{sameAs}(X_0,X_1)\ \land\ \dots\ \land\  \textit{sameAs}(X_{N-1},X_N)\ \land\ X_0<X_N.\\
\label{sameAs:complem_rule3}\tag{$3$}
& \textit{sameComp}(X,Y) \gets \textit{sameAs}(X,Y)\ \land\ \naf \textit{noStart}(X)\ \land\ X<Y.\\
\label{sameAs:complem_rule4}\tag{$4$}
& \textit{sameComp}(X,Z) \gets \textit{sameComp}(X,Y)\ \land\ \textit{sameAs}(Y,Z)\ \land\ X < Y\ \land\  X < Z.\\
\label{sameAs:complem_rule5}\tag{$5$}
& \textit{sameComp}(X,X) \gets \textit{sameComp}(X,Y).
\end{align}
Here, rule~(\ref{sameAs:complem_rule1}) is the symmetric closure of the {\em sameAs} predicate, representing the pairs of equivalent individuals.
Rules~(\ref{sameAs:complem_ruleh1})-(\ref{sameAs:complem_rulehn}) map into the unary predicate {\em noStart} all individuals that are greater than another individual at a distance less than or equal to $N$ with respect to the {\em sameAs} predicate.
Note that by $N = 0$ we mean that rules~(\ref{sameAs:complem_ruleh1})-(\ref{sameAs:complem_rulehn}) are not considered and \textit{noStart} is not populated.
Given a constant $c$ that is not part of the extension of {\em noStart}, rules~(\ref{sameAs:complem_rule3}) and~(\ref{sameAs:complem_rule4}) compute the pairs $(c,d)$ where $c < d$ and $d$ can be reached from $c$ via the {\em sameAs} transitive closure.
Since the minimum constant $c_{\text{min}}$ of every {\tt owl:sameAs}-clique $C$ can not populate the {\em noStart} predicate, it turns out that a fact {\em sameComp}$(c_{\text{min}},d)$ is generated for each constant of $C$.
Eventually, rule~(\ref{sameAs:complem_rule5}) is the reflexive closure of {\em sameComp} with respect to its first argument.

Even though rules~(\ref{sameAs:complem_rule1})-(\ref{sameAs:complem_rule5}) ensure, for any $N\geq 0$, that each element of a {\tt owl:sameAs}-clique is linked to the minimum of that clique by means of the {\em sameComp} predicate, many {\em sameComp$(c,d)$} facts could arise, where $c$ is not the minimum of any {\tt owl:sameAs}-clique.
The purpose of rules~(\ref{sameAs:complem_ruleh1})-(\ref{sameAs:complem_rulehn}) is precisely to avoid the generation of these extra facts as much as possible.
As we will see later, the {\it sameComp} predicate is used whenever any join variable occurs in a rule or query in order to enable matches between individuals belonging to the same clique, therefore it is very important to keep its growth under control.
It is easy to foresee that the larger $N$ is chosen the more the extension size of {\em sameComp} is reduced (due to the growing size of the extension of {\it noStart}),  but at the expense of a possible greater consumption of time due to the computation of the paths in the generation of {\it noStart} instances.
The choice of $N$ must therefore be weighted, according to the application domain at hand, in such a way that ideally, both the time and the extension size of {\it sameComp} are minimized.
More detailed considerations are reported in Section~\ref{sec:eval} where we considered different values of $N$ for the real-world domain of DBpedia and experimentally identified $N=2$ as the best compromise in the trade-off between time and space. %

Further attempts have been made to manage {\tt owl:sameAs} via Datalog (starting from the most naive one in which the reflexive, symmetrical and transitive closure is performed over the {\it sameAs} predicate directly, passing through more refined techniques in which the transitive closure relies on a fresh predicate name) but none of them was found to be applicable in practice.
This reinforces our approach and highlights how important is handling {\tt owl:sameAs} in an optimal way.

In the following we specify how we rewrite any rule in which appears at least a join variable.
For each rule $r$, let $J_r$ (the join variables of $r$) be the set of the variables occurring more than once in the body of $r$, and $2^{J_r}$ be the power set of $J_r$. For $X\in J_r$ we denote with $\#_X^r$ the number of occurrences of the variable $X$ in the body of $r$. For each $V \in 2^{J_r}$ we produce a new rule $r_V$ obtained from $r$ as follows: for any $X\in V$ and $1 \leq i\leq \#_X^r$, we substitute the $i$-th occurrence of $X$ with a fresh variable $X_{i}$ and add a new body literal $\text{sameComp}(X,X_i)$. For instance, with the rule in Example~\ref{example:griffin} we have that $J_r = \{Y\}$ and we get the following additional rule in the ontology:
    \begin{equation*}
        \text{DogOwner}(X) \gets \text{hasPet}(X,Y_1) \ \land\ \text{Dog}(Y_2) \ \land\ \text{sameComp}(Y,Y_1) \ \land\ \text{sameComp}(Y,Y_2).
    \end{equation*}
Intuitively, since in general the {\em sameAs} (and consequently {\em sameComp}) predicate does not involve all the constants of a program, any possible combination in which some join variables enables matches between constants syntactically different but linked by the {\tt owl:sameAs} relationship has to be considered.
We observe that, in most cases, only a few join variables occur in programs deriving from web ontologies.
However, in order to prevent a potentially exponential growth of the program, we rewrite rules that have more than $3$ join variables by projecting the {\em sameComp} predicate on each argument in which appears such a variable.
By doing this we avoid all the possible combinations of the variables and we get a single new rule that binds {\em sameComp} to all the join variables simultaneously.
%

\section{Related work}\label{sec:related}
In this section we mention the tools in the literature supporting OWL 2 RL reasoning, whose approach is to express inference tasks for OWL 2 in terms of inference tasks for Datalog.
Orel~\cite{DBLP:conf/dlog/KrotzschMR10} is a reasoning system which subsumes both the EL and the RL profile of the OWL 2 ontology language and its approach is based on a bottom-up materialisation of consequences in a database.
In particular, ontological information are stored as facts, whereas logical ramifications are governed by ``meta-rules'' that resemble the rules of a deduction calculus.
However, Orel supports neither SPARQL nor conjunctive queries.
OwlOntDB~\cite{DBLP:conf/fhies/FaruquiM12} works under the unique name assumption (UNA) to translate OWL 2 RL ontologies into Datalog programs, but it is no longer available.
DReW~\cite{DBLP:conf/csemws/XiaoEH12} is a query answering system which supports OWL 2 RL and OWL 2 EL (modulo datatypes). It uses \dlv as underlying Datalog engine and has not been conceived to generate ontology rewritings.
RDFox~\cite{DBLP:conf/semweb/NenovPMHWB15} is a main-memory RDF store supporting Datalog reasoning with an efficient handling of {\tt owl:sameAs} and SPARQL.
After the initial development at University of Oxford, the system is now available commercially from Oxford Semantic Technologies, a spin-out of the University backed by Samsung Ventures and Oxford Sciences Innovation.
Among systems aforementioned, only RDFox handles {\tt owl:sameAs}, whereas datatypes are supported by OwlOntDB and RDFox.
Clipper~\cite{DBLP:conf/aaai/EiterOSTX12} is a reasoner for conjunctive query answering over \hornshiq\ ontology.
Being more oriented to DL languages rather than OWL ontologies, Clipper lacks the support of the {\em datatypes} constructs and manages the {\tt owl:sameAs} property under the UNA.
It can also be used to generate ontology rewritings only.
\owltodlv~\cite{DBLP:conf/datalog/AlloccaCCCCFFGL19} is a commercial system, builts on the ASP reasoner \dlvdueserver~\cite{dlv2server}, for evaluating SPARQL queries over very large OWL 2 knowledge bases whose associated DL falls within \hornshiq. As well as Clipper, \owltodlv works under the UNA, but it cannot be used to generate ontology rewritings only.
\textsc{MASTRO}~\cite{DBLP:journals/semweb/CalvaneseGLLPRRRS11} and Ontop~\cite{DBLP:journals/semweb/CalvaneseCKKLRR17} are open-source tools for Ontology-Based Data Access (OBDA) in which the ontology lies
in the QL fragment of OWL 2. Ontop has its roots in \textsc{MASTRO} and is implemented through a query rewriting technique which avoids materializing triples.
$\dlv^\exists$~\cite{DBLP:journals/tocl/LeoneMTV19} is an effective system for fast query answering over {\tt shy}, an easily recognizable class of strongly parsimonious programs that strictly generalizes Datalog while preserving its complexity even under conjunctive queries. Among the systems dedicated to ontological query answering in the context of existential rules there are also Graal~\cite{DBLP:conf/ruleml/BagetLMRS15} and VLog~\cite{DBLP:conf/semweb/CarralDGJKU19}. These systems are not ad-hoc for OWL 2 RL and typically capture more expressive ontologies and then support features such as the {\em skolem} and the {\em restricted (standard) chase} for reasoning over existential rules.

\section{Experimental evaluation}\label{sec:eval}
To demonstrate the practical applicability of \sysname,
we designed and conducted an experimental evaluation based on the following working hypotheses: $(i)$ over synthetic OWL 2 RL benchmarks, {\em DaRLing}'s output is comparable with the one produced by existing tools in terms of both number of produced rules and quality of the rewriting (the latter measured via execution time fixed the Datalog engine); $(ii)$ over real-world OWL 2 RL knowledge bases, {\em DaRLing}'s rewriting strategy enables scalable query answering even in case the UNA is not a viable option.

\subsection{Set-up}
Since \sysname is a rewriter which does not rely on any Datalog engine, according to Table~\ref{tab:systems_features} and Section~\ref{sec:related}, the only tool that can be fairly tested against \sysname is the part of Clipper providing the Datalog rewriting of an OMQ, hereinafter called {\em Clipper-Rew}.
Moreover, since these rewriters are independent from the evaluation phase, for the purpose of our testing, the choice of the Datalog engine is immaterial. In our case, we simply opted for \idlv\footnote{See \url{https://github.com/DeMaCS-UNICAL/I-DLV}}~\cite{DBLP:conf/aiia/CalimeriFPZ16}, an Answer Set Programming instantiator and full-fledged state-of-the-art Datalog reasoner. Indeed, the system when fed with a disjunction-free and stratified under negation program is able to fully evaluate it and compute its perfect model.
As benchmarks, we relied on LUBM~\cite{DBLP:journals/ws/GuoPH05}, Adolena, Stock Exchange, Vicodì\footnote{See \url{http://www.vicodi.org}} and DBpedia~\cite{DBLP:conf/semweb/AuerBKLCI07}.

LUBM is an \hornshiq\ ontology over a university domain with synthetic data and $14$ queries. We restricted to the fragment falling in RL and considered its $14$ canonical queries.
Adolena, Stock Exchange and Vicodì have been derived from a well-established benchmark~\cite{DBLP:journals/tocl/LeoneMTV19}. They are expressed in DL-Lite$_R$ and provided with $5$ queries each. Adolena (Abilities and Disabilities OntoLogy for ENhancing Accessibility) has been
developed for the South African National Accessibility Portal. Stock Exchange is an ontology of the domain of financial institution within the EU.
Vicodì is an ontology of European history.
For LUBM we adopted $6$ datasets, wherease for each of Adolena, Stock Exchange and Vicodì, we used $5$ datasets, downloaded from \url{https://www.mat.unical.it/dlve}.

Differently from the other aforementioned benchmarks, that are synthetic and assume the UNA, DBpedia is a real-world knowledge base requiring a proper handling of the {\tt owl:sameAs} property. More in detail, DBpedia is a well-known KB falling in OWL 2 RL and created with the aim of sharing on the Web the multilingual knowledge collected by Wikimedia projects in a machine-readable format. The dataset has been extracted from the latest stable release of the whole DBpedia dataset. The considered part consists of the English edition of Wikipedia and is composed by about half a billion RDF triples. We inherited a set of $10$ queries from an application conceived to query DBpedia in natural language~\cite{dlv2server}.
%
All tests were performed on a machine having two 2.8GHz AMD Opteron 6320 processors and $128$ GB of RAM.
All rewritings for each benchmark and executables are available at \url{https://demacs-unical.github.io/DaRLing}.

\begin{figure}[t]
    \centering
    \includegraphics[width=0.84\textwidth]{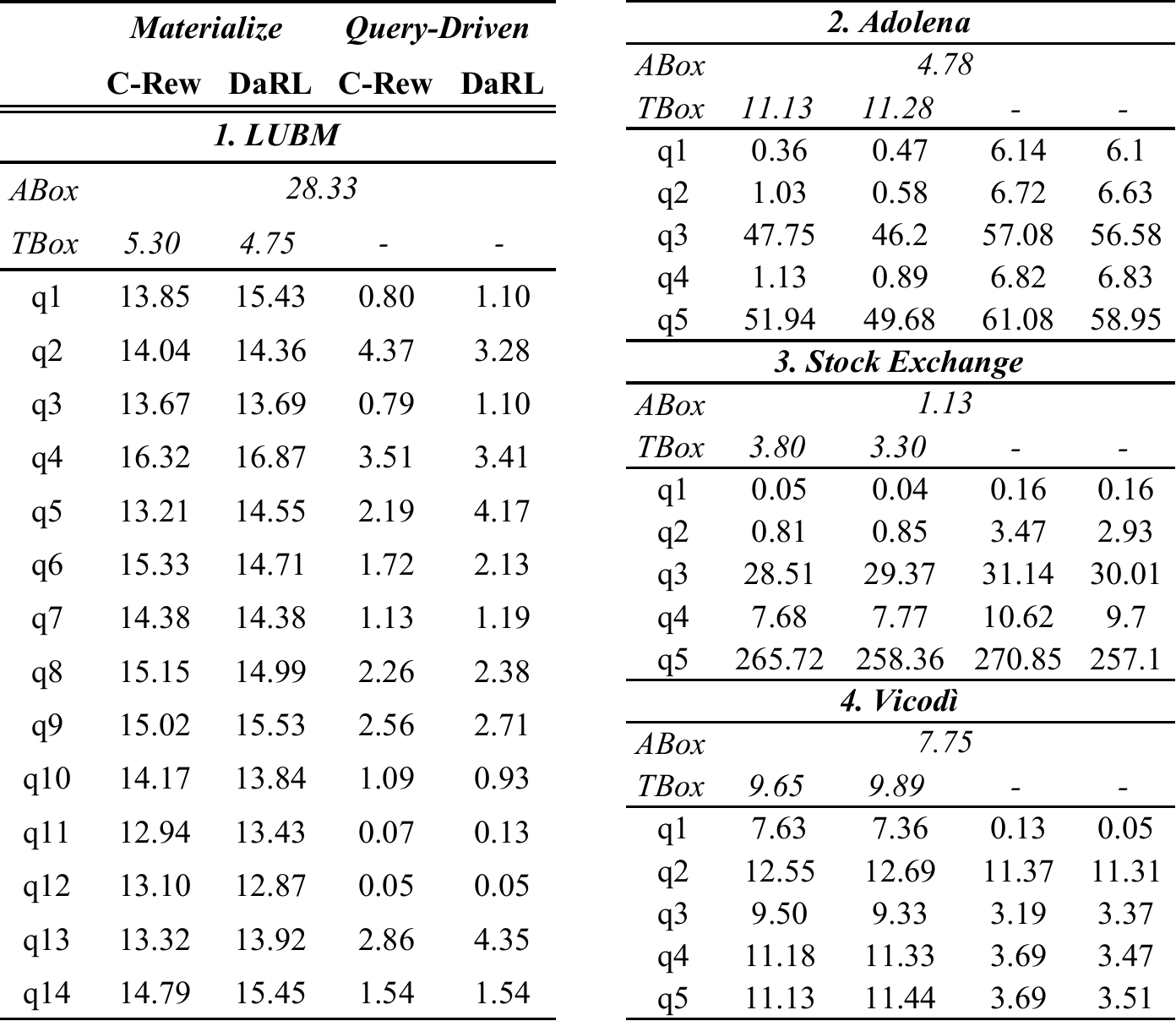}
    \caption{Experiments on LUBM, Adolena, Stock Exchange and Vicodì. C-Rew stands for Clipper-Rew, DaRL for DaRLing. Times are in seconds.}
    \label{tab:lubm-vicodi}
    \vspace{-1\baselineskip}
\end{figure}

\subsection{Quality}
In this former set of experiments, we first generated {\em Clipper-Rew} and \sysname rewritings for all queries of LUBM, Adolena, Stock Exchange and Vicodì; then, over these rewritings and for all considered datasets, we executed \idlv under two different scenarios: in the scenario {\em materialize} the system is forced to materialize the whole ontology and then prompted to answer to each query individually; in the scenario {\em query-driven} the system still runs each query one by one, but performs a more efficient evaluation tailored on the query at hand by enabling the magic sets technique~\cite{DBLP:journals/tplp/AlvianoLVZ19}.

Figure~\ref{tab:lubm-vicodi} reports average running times in seconds of \idlv executions over all datasets on LUBM, Adolena, Stock Exchange and Vicodì, respectively. In particular, for better highlighting differences in performance, for each execution we extracted from total time the static and fixed times spent on ABox loading and on TBox materialization over the ABox. These fixed times are reported in the table as {\em ABox} and {\em TBox}, respectively.
Notice that TBox materializing times are not reported in case of scenario {\em query-driven} as there no materialization is done.
Results show that in most cases performance achieved by \idlv when using \sysname outputs is comparable w.r.t. {\em Clipper-Rew}. Some worsening is observable especially on LUBM queries $5$ and $13$. This is reasonable since {\em Clipper-Rew} requires to take into account the query at hand for properly translating the query and the TBox, thus generating an output optimized on the basis of the query. \sysname instead follows a different principle as it is designed to produce general and query-independent TBox rewritings without specific query-oriented enhancements.
On Adolena and Stock Exchange, \idlv times with \sysname rewritings are generally, slightly better than with {\em Clipper-Rew}.
Regarding Vicodì, \idlv performance are practically the same since {\em Clipper-Rew} and \sysname produced almost identical rewritings. This is mainly because Vicodì TBox, when rewritten by both \sysname and {\em Clipper-Rew} into Datalog, consists of linear rules, i.e., rules having only an atom in body; thus, when the presence of joins is limited both approaches result almost equivalent.
%

\subsection{Scalability}

We also experimented on DBpedia with two types of rewritings of $10$ DBpedia queries generated via \sysname.
In particular, we first measured the costs, in terms of both time and space, of materializing the {\it sameComp} predicate by varying the value of $N$ (see Section~\ref{sec:sameAs}).
To this end, we generated via \idlv the materialization of the whole TBox under the UNA and filtered out the tuples of the {\it sameAs} predicate. The resulting dataset has been paired with rules~(\ref{sameAs:complem_rule1})-(\ref{sameAs:complem_rule5}) by considering different values of $N$.
Table~\ref{table:tradeoff_sameAs} reports the values of the following three parameters by varying $N$: the time (in seconds) needed for materializing the {\it sameComp} predicate, the extension size of the {\it sameComp} predicate and the memory consumption in GB.

\begin{table}[t]
\caption{Costs of the DBpedia {\it sameComp} materialization (time limit set to $10$ hours).}
\label{table:tradeoff_sameAs}
  \begin{minipage}{\textwidth}
		\centering
\begin{tabular}{lccccc}
\hline\hline
{Parameter} & {N = 0}           & {N = 1}       & {N = 2}      & {N = 3}      & {N = 4}    \\
\hline
\textit{Time (seconds)} & $5,033$             & $2,673$      & $864$     & ~~~$26,355$~~~     & ~~~timeout~~~   \\
\textit{\#sameComp} & ~~~$523$M~~~           & ~~~$342$M~~~ & ~~~$103$M~~~ & $77$M & -  \\
\textit{Memory (GB)} & $35$ & $23$        & $7$        & $5$        & -   \\
\hline\hline
\end{tabular}
\vspace{-1\baselineskip}
\end{minipage}
\end{table}

Ideally, the optimal value of $N$ should be the one minimizing both the time and the space (the memory consumption or, equivalently, the extension size of {\it sameComp}). As it is evident from Table~\ref{table:tradeoff_sameAs}, such a unique value does not exist. Indeed,
the minimum value of time is reached for $N = 2$, whereas the minimum value of space is reached for $N=3$. Between the two values, the best compromise seems to be offered by $N=2$, since a small blow-up in terms of space is highly reward in terms of time saving.
\begin{figure}[b]
  \centering
    \includegraphics[width=0.6\textwidth]{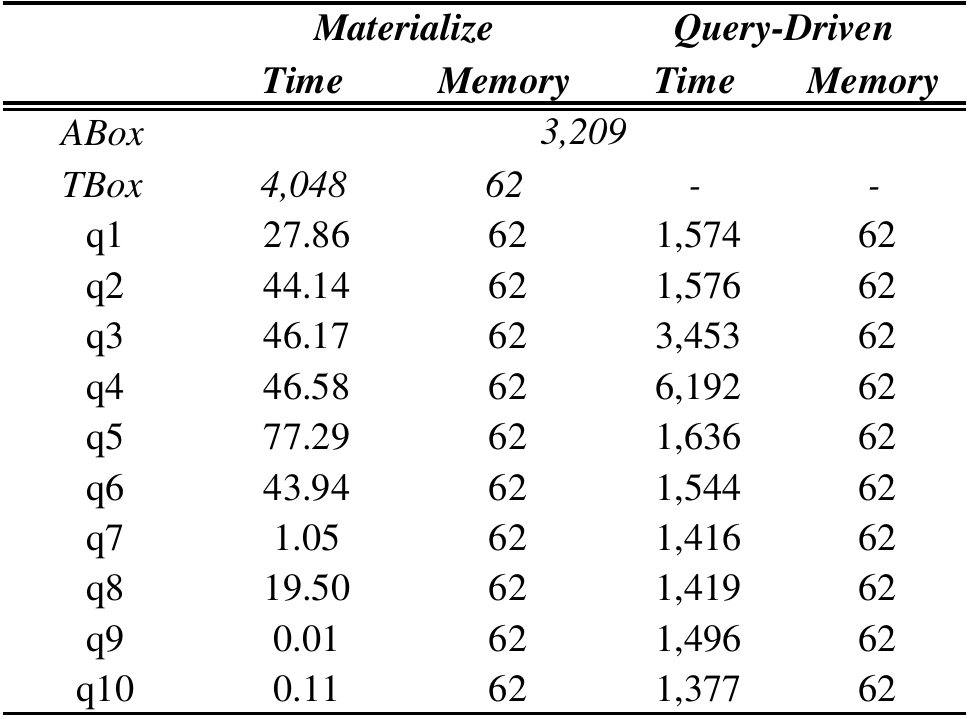}
   \caption{Experiments on DBpedia for $N=2$. Times are in seconds, memory is in GB.}
   \label{tab:dbpedia}
   \vspace{-1\baselineskip}
\end{figure}%

We thus decided to focus on $N=2$ and to evaluate, per each of the $10$ queries, performance when \idlv is provided with \sysname rewritings generated for this value of $N$.
Results are reported in Figure~\ref{tab:dbpedia}.
We restricted experiments to \sysname as we purposely want to investigate scalability of \sysname under the non-UNA. In addition, comparisons with other tools (cfr. Table~\ref{tab:systems_features}) would result unfair since to our knowledge, \sysname is the only open-source project empowered with an ad-hoc handling of the {\tt owl:sameAs} property.
As in quality-measuring experiments, we considered both the {\em materialize} and {\em query-driven} scenarios; times are in seconds and memory in GB. {\em ABox} and {\em TBox} times, as in the above results, represent times spent on ABox loading and on TBox materialization over the ABox, respectively. Columns report for each query running times from which {\em ABox} and {\em TBox} times have been subtracted. In the {\em materialize} scenario, we observe that in general, once the TBox is materialized, the system is able to compute query answers in at most $4$ minutes. A greater effort is paid if instead of materializing the TBox, the system is requested to perform a query-driven computation. In all cases, the total times spent on single queries are up to $4$ times smaller than the time spent for materializing just the {\em sameComp} predicate for $N=3$ as showed in Table~\ref{table:tradeoff_sameAs}. This behaviour reinforces the choice of $N=2$ as the best value for this domain. Concerning memory, we can conclude that in both scenarios, despite the large ABox, \sysname rewritings permit to \idlv to evaluate all queries within $62$ GB.
%

%

\section{Conclusions}\label{sec:conclusion}
In this paper we have presented {\em DaRLing},
a Datalog-based rewriter for OWL 2 RL ontological reasoning under SPARQL queries.
To demonstrate its practical applicability,
we have designed and conducted an experimental evaluation based on two working hypotheses, which have been confirmed. The first release of \sysname demonstrates to produce more general rewritings equivalent or sightly differing from the ones generated by Clipper, the closer open-source competitor of \sysname. As additional feature, \sysname can be used for transparently handling the {\tt owl:sameAs} property independently from the Datalog reasoner at hand albeit requiring extra work due to the intrinsic need of computing the transitive closure. Such costs are strictly dependent from the ontology at hand; our experimentation in an unfriendly setting of a large ontology such as DBpedia proved a not taken for granted applicability of the approach.

Concerning our future plans, the ultimate objective is to develop a freely available customizable SPARQL endpoint for OWL 2 RL ontological reasoning complying with the W3C Recommendations.
To this end, the next steps are:
$(i)$ extending the rewriting to enable the meta-reasoning, namely SPARQL queries where variables may range also over the given schema;
$(ii)$ handling {\tt owl:sameAs} also over concepts and roles;
$(iii)$ enriching the set of supported datatypes;
$(iv)$ specializing the rewriting of the TBox by taking into account also the query;
$(v)$ specializing the rewriting of the query by taking into account also the TBox.

\bibliographystyle{acmtrans}
\newcommand{\SortNoOp}[1]{}

\label{lastpage}
\end{document}